\def\year{2021}\relax
\documentclass[letterpaper]{article} 
\usepackage{aaai21}  
\usepackage{times}  
\usepackage{helvet} 
\usepackage{courier}  
\usepackage[hyphens]{url}  
\usepackage{graphicx} 
\usepackage{amsfonts,amssymb}
\usepackage{amsmath}
\usepackage{color}
\usepackage{bm}
\usepackage{multirow}
\usepackage{booktabs}
\usepackage[switch]{lineno}
\usepackage{natbib}  
\usepackage{caption} 
\title{RTS3D: Real-time Stereo 3D Detection
	from 4D Feature-Consistency Embedding Space for Autonomous Driving}
\author{
   Peixuan Li,
   Shun Su,
   Huaici Zhao\\
}
\affiliations{
    University of Chinese Academy of Sciences\\

}
\begin{document}

\maketitle

\begin{abstract}
	Although the recent image-based 3D object detection methods using Pseudo-LiDAR representation have shown great capabilities, a notable gap in efficiency and accuracy still exist compared with LiDAR-based methods. Besides, over-reliance on the stand-alone depth estimator, requiring a large number of pixel-wise annotations in the training stage and more computation in the inferencing stage, limits the scaling application in the real world.
	In this paper, we propose an efficient and accurate 3D object detection method from stereo images, named RTS3D. Different from the 3D occupancy space in the Pseudo-LiDAR similar methods, we design a novel 4D feature-consistent embedding (FCE) space as the intermediate representation of the 3D scene without depth supervision. The FCE space encodes the object's structural and semantic information by exploring the multi-scale feature consistency warped from stereo pair. Furthermore, a semantic-guided RBF (Radial Basis Function) and a structure-aware attention module are devised to reduce the influence of FCE space noise without instance mask supervision. Experiments on the KITTI benchmark show that RTS3D is the first true real-time system (FPS$>$24) for stereo image 3D detection meanwhile achieves $10\%$ improvement in average precision comparing with the previous state-of-the-art method.
The code will be available at \url{https://github.com/Banconxuan/RTS3D}
\end{abstract}
\section{Introduction}
\begin{figure}[!htb]
	\begin{center}
		\includegraphics[width=0.9\columnwidth]{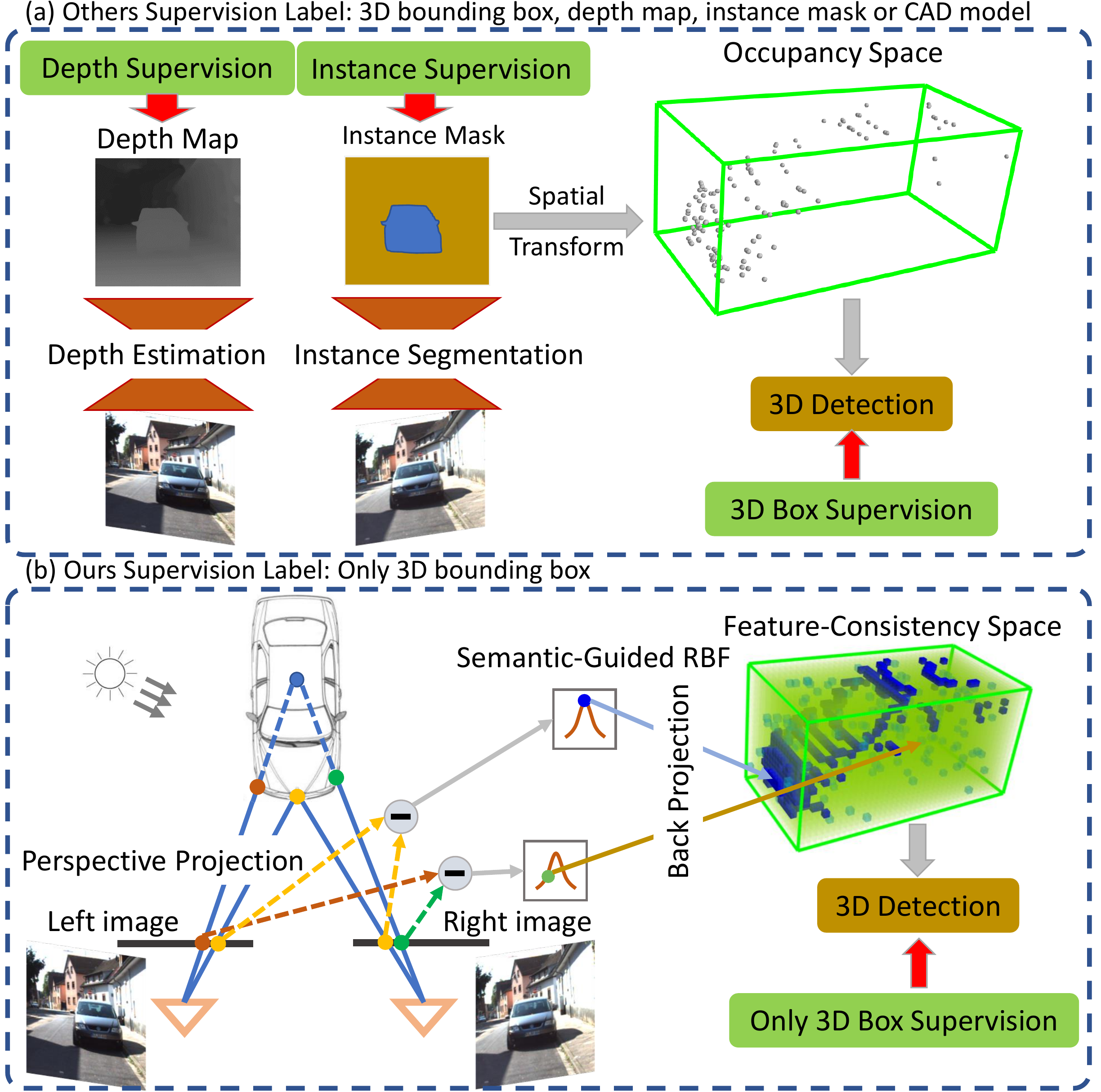}
	\end{center}
	\caption{Comparisons between occupancy space and proposed FCE space: (a) execute a depth generator to encode the structure of the object in a 3D occupancy space, and use instance mask to reduce the influence of noise in non-target areas. By contrast, our proposed RTS3D, as shown in (b), encode the structure of the object by estimate the consistency between warped left and right images for each 3D locations, and explicitly model the semantic cues for noise filtering, yielding superior accuracy and efficiency without additional label supervision. }
	\label{fig:abs}
\end{figure}
3D object detection serves as an important role in many applications, such as augmented reality, robotics, and autonomous driving.
Although recently developed LiDAR-based detection algorithms \cite{conf/cvpr/ShiGJ0SWL20,conf/cvpr/HeZH0Z20} show some excellent performance, the high price, low service life, and discordant appearance of the LiDAR system restrict its further development in practical applications. Alternatively, the solutions relying on cameras are very competitive for its low-cost, low-power consumption, and high flexibility in development.  Therefore, it is grabbing much more attention in computer communities recently \cite{conf/cvpr/Chen0SJ20,conf/cvpr/SunCXZJZB20,conf/cvpr/0001CS19}.

Image-based methods have two main tasks: 1) to find an appropriate and effective representation to recover the geometric structure of a 3D scene, and  2) to eliminate the interference of non-target areas. For the first task, Wang et al. have proposed a Pseudo-LiDAR representation \cite{wang2019pseudo}, which expand 3D object detection from the 2D frontal view space to the 3D occupancy space. Recent methods try to tackle the second task by designing an instance-level Pseudo-LiDAR generator \cite{journals/corr/abs-1909-07566,journals/corr/abs-2003-00529,conf/cvpr/SunCXZJZB20}, which only estimate the depth map of the objects of interest. However, all these approaches heavily rely on extra sub-networks to perform CAD model generation \cite{conf/cvpr/SunCXZJZB20}, instance segmentation \cite{journals/corr/abs-2003-00529} or depth map estimation \cite{conf/cvpr/SunCXZJZB20,wang2019pseudo,journals/corr/abs-2003-00529}, as shown in Fig. \ref{fig:abs} a. The additional pixel-wise labels for supervised learning required in these sub-networks become the biggest obstacle in collecting labor-intensive annotations, make it impractical in many real application scenarios. Moreover, reliance on stand-alone sub-networks makes an inherent disconnection in transmitting gradient while consuming plenty of computing resources in the training and inferring stages, limiting upper-bound of the detection accuracy and speed. Here, we tackle these two tasks without relying on additional labels while achieving true real-time detection with competitive accuracy against the state-of-the-art method.

The main contribution of our approach is a novel 4D intermediate representation of 3D object structure, named FCE space. This is different from the previous 3D occupancy space in Pseudo-LiDAR similar methods that represent object structure by estimating whether a location is occupied or not, as shown in Fig. \ref{fig:abs} a). Here, we encode the structure of underlying objects by the feature consistency between warped left and right images for each 3D locations in latent space, as shown in Fig. \ref{fig:abs} b.  The rationalization behind the proposed representation comes from a typical assumption that the intensity of light projected onto the stereo image from the visible surface of a 3D object should be more consistent than from the non-object surface. The same assumption is also used in the plane-sweeping method \cite{conf/cvpr/Collins96} to estimate the depth map, thus proving that the consistency space can encode structural information.    We aim to establish such a consistency space and directly detect objects on it.

However, establishing FCE space is complex in the computation of the entire camera visual range, and this unsupervised space contains an enormous amount of noise due to the interference of non-Lambert properties, the textureless region, and nontarget surface.

We address these issues in four steps. First, we only compute feature consistency in the latent space of the target object. The initial latent space is predicted by monocular 3D detection at a high speed. Later it would be iteratively refined by the detection results of FCE space.
Second, we compute the consistency from the multi-scale feature to make it more reliable in textureless and reflective regions. Predicting the required consistency only need local neighborhood, so a very simple convolutional neural network(e.g. ResNet18 \cite{he2016deep}) is adopted to extract the multi-scale features. Third, we propose to encode the semantic information in an RBF to reduce the interference of the nontarget surface. This semantic-guided RBF explicitly modeling the semantic cues to 3D space is easier to converge than implicit learning possible relationships. Fourth, we propose a structure-aware attention (StrAA) module to further filter the spatial noise and capture local structure at a smaller computational cost than 3D CNN and PointNet++\cite{conf/nips/QiYSG17}.

To summarize, Our contributions are as follows:
	\textbf{1.)} An image-based 3D object detection approach predicts the 3D box of objects more efficiently and accurately.
	\textbf{2.)} A novel intermediate representation of object structure that bridges the performance gap between LiDAR-based and image-based methods without additional label supervision.
	\textbf{3.)} A semantic-guided RBF and a StrAA module to reduce the interference of noise and optimize the characterization of local
	structure in the FCE space.
	\textbf{4.)} Evaluation on the popular KITTI dataset shows that the proposed method is the first true real-time 3D detection approach using only images and achieves comparable detection accuracy against the other competitors.
\section{Related Work}
\begin{figure*}[!htb]
	\begin{center}
		\includegraphics[width=1.9\columnwidth]{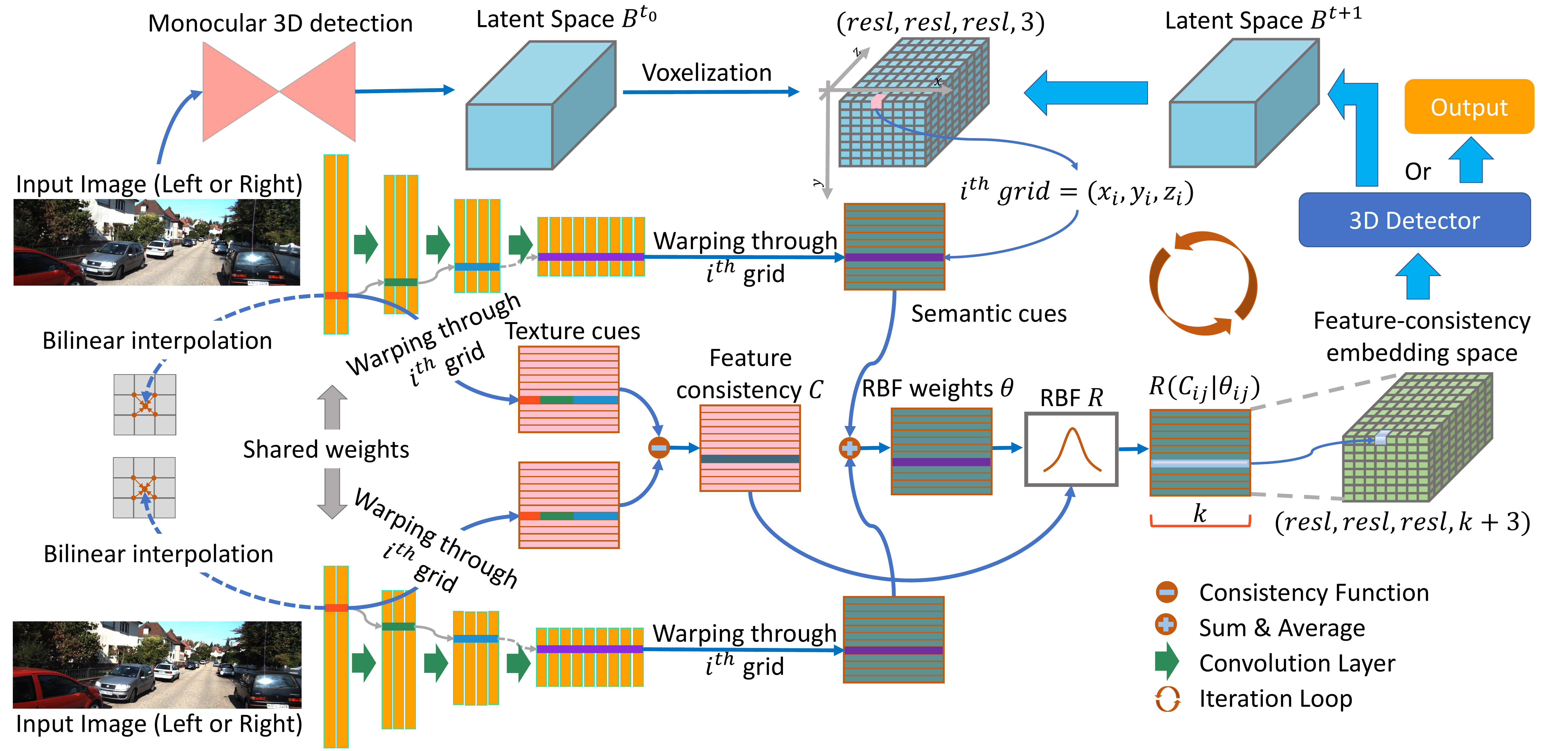}
	\end{center}
	\caption{Overview of RTS3D architecture. Stereo images are first passed through a simple siamese network to generate the multi-scale feature. In parallel, a coarse latent space is predicted by fast monocular 3D detection and then is split to a regular grid. The FCE space is generated by warping the left and right multi-scale features to each location in latent space after the semantic-guided RBF. The 3D detector estimates the 3D box from the FCE space as the final output or generates a more refined hidden space for the next iteration.}
	\label{fig:framework}
\end{figure*}
\textbf{Monocular 3D Object Detection.} Due to the lack of depth, 3D object detection is difficult given only a monocular image. A common theme of these methods is to employ sub-networks to generate extra 2.5D feature, such as depth map\cite{xu2018multi,ma2019accurate}, object mask\cite{chen2016monocular}, or CAD model\cite{chabot2017deep}. Recent monocular-only works attempt to apply the geometry constrain as the post-processing \cite{mousavian20173d,li2020rtm3d,li2020monocular} or embedding knowledge to aid in detection. These methods explicitly model the relationship between 3D location and 2D feature, which enables them to be improved in both accuracy and running speed. However, their promised accuracy still not good enough comparing stereo approaches.\\
\textbf{Stereo-based 3D Object Detection.}
Like monocular approaches, stereo methods can also be roughly divided into two ways by the type of training data. One is Pseudo-LiDAR similar pipeline. These methods \cite{wang2019pseudo,journals/corr/abs-1906-06310,conf/cvpr/Chen0SJ20} first use a SOTA disparity prediction with stereo processing to generate a depth map following to convert this depth map to occupancy space. Then apply a LiDAR-based framework \cite{conf/cvpr/ShiWL19,qi2017pointnet} to detect object. In order to save computation and avoid streaking noise caused by non-target regions, the recent method aims to detect object only in potential area by introducing the instance mask \cite{conf/cvpr/SunCXZJZB20,journals/corr/abs-1909-07566,journals/corr/abs-2003-00529,CSCL_Dong_ECCV2020}. Intuitively, these methods containing more prior information, from extra-label supervision, would certainly improve the performance of detection. However, reliance on additional sub-networks and labels also leads to more time consumption and labor-intensive work. Another one, therefore, tries to fully explore the potency of stereo images. Stereo R-CNN \cite{conf/cvpr/0001CS19} associate left and right 2D box to generate rough 3D box that are later refined by dense 3D box alignment. TLNet \cite{journals/corr/abs-1906-01193} enumerate a multitude of 3D anchors and then construct object-level correspondences to filter out dreadful proposals. However, these methods perform a lower accuracy and running speed comparing with the Pseudo-liDAR similar methods. By comparison, the proposed method has the fastest running and achieve a competitive accuracy comparing with the Pseudo-liDAR without extra labels help.

\section{Proposed Method}
Given a stereo pair $(I_L, I_R)$, the goal is to estimate the 3D property of object typical represented by $B=(X, Y, Z, W, H, L, \theta)$, which denotes the 3D center position, width, height, length, and horizontal orientation respectively. Fig. \ref{fig:framework} shows an overview of the proposed framework. It comprises four stages: 1) A very fast monocular 3D detector is leveraged to obtain initial latent space. 2) Multi-scale features are back-projected onto the grid of initial latent space to construct FCE space 3) A semantic-guided RBF and structure-aware attention module reduce the influence of FCE space noise and optimize the characterization of local structure. 4) A variant of the PointNet to predict the final 3D box with its confidence or generate more specific latent space for the next iteration.

\subsubsection{Latent Space Generation.}
Instead of creating an entire viewable FCE space, we only computer the feature consistency in latent space containing the object of interest. 
Benefiting from the recent development of monocular 3D detection, we propose to employ an efficient one of them to generate an initial coarse cuboid $B^{t_0}$ as the guidance of latent space. Later this coarse cuboid can be iteratively refined as $B^{t+1}$ by the detection results of established FCE space.  
Here, we choose two monocular 3D object detection frameworks for the trade-off between speed and accuracy: KM3D-Net \cite{li2020monocular} and CenterNet \cite{zhou2019objects}. Both structures are one-stage 3D detectors and do not rely on extra annotation for the training.
\subsubsection{Multi-scale Texture Cues Generation.}
Inspired by traditional stereo matching methods \cite{conf/cvpr/ZhangFMSYYT14}, which process the correspondence by texture cues across multiple scales, we generate the consistency of a 3D location from pair images by extracting the hierarchical contextual information of low-level features. To ensure the real-time performance,the simple convolutional encoding structures, ResNet18\cite{he2016deep}, is adopted to output the multi-feature $\{F^s_{l\,r}\}_s^S$ with the downsampling stride $s={2, 4, 8}$. However, relying heavily on texture cues will unavoidably introduce noise from non-target objects, such as the ground or other objects. To overcome this issue, we add one high-level feature output with downsampling stride $/32$ to predict semantic cues. Nevertheless, without the instance mask and depth map annotation supervision, implicit fusion of this information makes the model difficult to converge. We, therefore, design a semantic-guided RBF to explicitly encode two cues for noise filtering.
\subsubsection{Building the Feature-Consistency Embedding Space.}\label{section:buildFCES}
Given the latent space, texture cues and semantic cues on object of interest, we convert them to the FCE space to encode geometric structure. We first split the latent space to regular grid with resolution ratio $resl$, which represent the latent space as $G=\{g_i=[x_i,y_i,z_i]\in \mathbb{R}^3\}_{i=1...resl \times resl \times resl}$. After that, we project a voxel $g_i$ into feature space $x_i^s=[u_i^s,v_i^s,1]^T$ by using camera intrinsics $K$, extrinsic parameters $T$, consisting of a rotation matrix $R$ and a translation matrix $t$ of the left and right camera, and coordinate affine transformation $h_s$ of the original image into the multi-scale features:
\begin{equation}
	\label{eq:a}
	\begin{aligned}
		\sideset{_{}^{l\,r}}{_{i}^{s}}{\mathop{x}} =h_s K_{l\,r}
		\left[
		\begin{matrix}
			R^{3 \times 3}_{l\,r}& t^{1 \times 3}_{l\,r}\\
			&\\
			0^T&1
		\end{matrix}
		\right]g_i
	\end{aligned}
\end{equation}
where $l\,r$ indicates it belongs to the left or right image. The purpose of introducing affine transformation matrix $h_s$ without uniform zooming parameters factor is to reduce the quantization error caused by different downsampling stride of original image scaling. Then the consistency of the 3D voxel $g_i$ from the left and right image can be defined as:
\begin{equation}
	\label{eq:a}
	\begin{aligned}
		C_i^s=f\left(\hat{F}_{l}(\sideset{_{}^{l}}{_{i}^{}}{\mathop{x}}),\hat{F}_{r}(\sideset{_{}^{r}}{_{i}^{}}{\mathop{x}})\right), \hat{F}_{l}=Cat\left[\hat{F}_{l}^{s}\cdots \right]_{s=2}^{8}
	\end{aligned}
\end{equation}
Here, $Cat$ means concatenation. Note that the projected coordinates $x_s^i$ are continuous values and the feature vectors are all integer coordinates. We, therefore, use the differentiable bilinear sampling mechanism $\hat{F}$ inspired by spatial transformer networks \cite{Jaderberg2015Spatial}. $f$ is a pair function of measures distance that represents the similarity of two signals. There are many existing choices for $f$, such as absolute difference, gaussian distance, cosine correlation, and concatenation. However, The first three methods are difficult to encode semantic cues and the uncertainty of each dimension. Concatenation implicitly encodes the uncertainty, but it is difficult to learn without the supervision of depth maps. We propose a novel semantic-guided RBF to explore pair signal relationship by combine texture cues and semantic cues:
\begin{equation}
	\label{eq:a}
	\begin{aligned}
		\hat{C}_i^s=RBF(\hat{F}_{l}(\sideset{_{}^{l}}{_{i}^{}}{\mathop{x}})-\hat{F}_{r}(\sideset{_{}^{r}}{_{i}^{}}{\mathop{x}})| \alpha_i)\\
	\end{aligned}
\end{equation}
$RBF$ denotes Radial Basis Function with parameters $\alpha$ that is normally the variance of a multi-scale feature in a given coordinate of voxel. Here, we consider $\alpha_i=\frac{1}{2}\left(\hat{F}^{32}_{l}(\sideset{_{}^{l}}{_{i}^{32}}{\mathop{x}})+\hat{F}^{32}_{r}(\sideset{_{}^{r}}{_{i}^{32}}{\mathop{x}})\right)$ in the form of learnable parameters from semantic cues. By doing this, $\alpha$ will be reduced in unreliable channels and non-target location in the image feature.
\subsubsection{3D Bounding Box Prediction.}
\begin{figure}[!htb]
	\begin{center}
		\includegraphics[width=1\columnwidth]{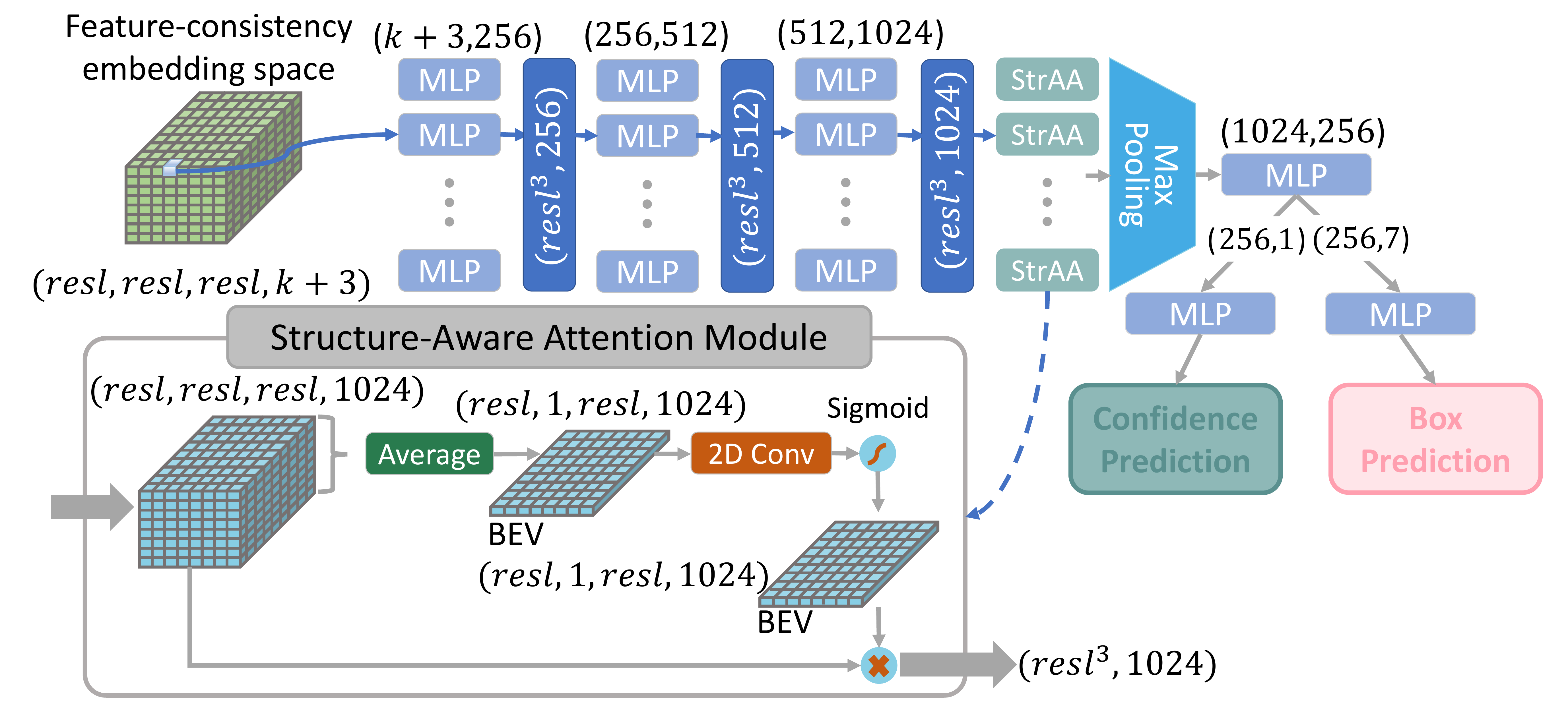}
	\end{center}
	\caption{Overview of the proposed 3D object detector. }
	\label{fig:box}
\end{figure}
After generating the FCE space in the grid form, the common solution is to employ 3D convolution networks(3DCNNs) to extract local features for the estimation of 3D bounding. However, we set a very small resolution (min $resl=10$ in our experiments) for the trade-off between accuracy and speed. This makes it difficult to determine the size of the 3D convolution kernel. 
For example, a large convolution kernel will introduce a lot of padding noise, while a small convolution kernel will increase computation but it is not obvious to extract local features. Here, we design a variant of Point-Net\cite{qi2016pointnet} with a StrAA module for 3D box prediction and confidence estimation, as shown in Fig. \ref{fig:box}\begin{table*}[htb]
	\caption{Comparison 3D detection methods for car category, evaluated by metric $AP_{3D}$ on \bm{$val$} / \bm{$test$} set on KITTI. Extra means the extra required data. \textcolor{blue}{Blue} indicates the highest result comparing other methods without extra labels. The value in parentheses indicates the gap between our results and the best results of other methods that require extra labels.}
	\begin{center}
		\resizebox{\textwidth}{!}{
			\begin{tabular}{| c  | c  | c | c | c | c | c@{/}c | c@{/}c | c@{/}c |}
				\hline
				\multirow{2}{*}{Method} & \multirow{2}{*}{Extra} & \multirow{2}{*}{Time} & \multicolumn{3}{c|}{$IoU>0.5$     [$ \bm{val$}]} & \multicolumn{6}{c|}{$IoU>0.7$ [\bm{$val/test$}]} \\
				\cline{4-12}
				& & & Easy & Moderate & Hard & \multicolumn{2}{c|}{Easy} & \multicolumn{2}{c|}{Moderate} & \multicolumn{2}{c|}{Hard} \\
				\hline
				3DOP \cite{chen20153d} & Instance Mask & - & 46.0 & 34.6  & 30.1  & 6.6 &- & 5.1 &-& 4.1 & -\\

				MLF \cite{xu2018multi} & Depth& - & - & 47.4 & -  & -&- & 9.8 &-& - & -\\
				RT3DStereo \cite{conf/itsc/KonigshofSS19}& Depth+Semantic Mask & \textbf{92ms}& - & -  & -  & - &28.5 & - &24.1& - & 20.32\\
				PL: F-PointNet \cite{wang2019pseudo} & Depth+Flow & 670ms & 89.5 & 75.5  & 66.3  & 59.4 &39.7 & 39.8 &26.7& 33.5 & 22.3\\
				PL: AVOD \cite{wang2019pseudo} & Depth+Flow & 510ms & 88.5 & 76.4  & 61.2  & 61.9 &55.40 & 45.3 &37.17&39.0& 31.37\\
				PL++: AVOD \cite{journals/corr/abs-1906-06310} & Depth+Flow & 500ms & 89.0 & 77.8  & 69.1  & 63.2 &- & 46.8 &-& 39.8 & -\\
				PL++: P-RCNN \cite{journals/corr/abs-1906-06310} & Depth+Flow & 510ms & 88.0 & 73.7  & 67.8 & 62.3 &- & 44.9 &-& 41.6 & -\\
				OC-Stereo \cite{journals/corr/abs-1909-07566} & Depth+Instance Mask & 350ms & 89.65 & \textbf{80.03} & 70.34  & 64.07 &55.15 & 48.34 &37.60& 40.39& 30.25\\
				
				ZoomNet \cite{journals/corr/abs-2003-00529} & Depth+Instance Mask & - & 90.44 & 79.82 & \textbf{70.47}  & 62.96 &55.98& \textbf{50.47}  &38.64& \textbf{43.63} & 30.97\\
				Disp R-CNN \cite{conf/cvpr/SunCXZJZB20} & Depth+Instance Mask+CAD & 425ms & \textbf{90.47} & 79.76   & 69.71 & \textbf{64.29} &\textbf{59.58} &47.73 &\textbf{39.34}& 40.11 & \textbf{31.99}\\				
				\hline
				\hline
				TL-Net \cite{conf/cvpr/0001CS19} & None & -& 59.51 & 43.71  & 37.99 & 18.15 &- &14.26 &-& 13.72 & - \\	
				Stereo RCNN \cite{conf/cvpr/0001CS19} & None & 417ms& 85.84& 66.28  & 57.24 & 54.11 &49.23 & 36.69 &34.05&31.07& 28.39\\				
				\hline
				Ours($iteration$=0, $resl=20$) & None & 30.2ms& 89.46 & 77.30  & 62.36  & 60.33 &54.12& 44.48 &34.59&37.99 & 28.91\\
				Ours($iteration$=1, $resl=20$) & None & \textbf{\textcolor{blue}{39.4ms}}&\textcolor{blue}{90.34}\textcolor{red}{(-0.13)} & \textcolor{blue}{79.67}\textcolor{red}{(-0.36)}  & \textcolor{blue}{70.29}\textcolor{red}{(-0.18)}  & \textcolor{blue}{64.76}\textcolor{green}{(+0.47)} &\textcolor{red}{58.51(-1.01)} & \textcolor{blue}{46.70}\textcolor{red}{(-3.77)} &\textcolor{blue}{37.38}\textcolor{red}{(-1.96)}&\textcolor{blue}{39.27}\textcolor{red}{(-4.36)}  & \textcolor{blue}{31.12}\textcolor{red}{(-0.87)} \\
				\hline
		\end{tabular}}
		\label{tab:3d}
	\end{center}
\end{table*}
\begin{table*}[htb]
	
	\caption{Comparison 3D detection methods for car category, evaluated by metric $AP_{BEV}$ on \bm{$val$} / \bm{$test$} set on KITTI. }
	\begin{center}
		\resizebox{\textwidth}{!}{
			\begin{tabular}{| c  | c  | c | c | c | c | c@{/}c | c@{/}c | c@{/}c |}
				\hline
				\multirow{2}{*}{Method} & \multirow{2}{*}{Accelerator} & \multirow{2}{*}{FPS} & \multicolumn{3}{c|}{$IoU>0.5$     [$ \bm{val$}]} & \multicolumn{6}{c|}{$IoU>0.7$ [\bm{$val/test$}]} \\
				\cline{4-12}
				& & & Easy & Moderate & Hard & \multicolumn{2}{c|}{Easy} & \multicolumn{2}{c|}{Moderate} & \multicolumn{2}{c|}{Hard} \\
				\hline
				3DOP \cite{chen20153d} & - & - & 55.0 & 41.3  & 34.6  & 12.6 &- & 9.5 &-& 7.6 & -\\
				MLF \cite{xu2018multi} & -& - & - & 53.7  & -  & - &- & 19.5 &-& - & -\\
				RT3DStereo \cite{conf/itsc/KonigshofSS19}& TITAN X & \textbf{11.0} & 25.19 & 18.20  & 15.52  & - &59.32& - &49.48& - & 43.16\\
				PL: F-PointNet \cite{wang2019pseudo} & - & 1.5 & 89.8 & 77.6 & 68.2 & 72.8 &55.0 & 51.8 &38.7& 44.0 & 32.9\\
				PL: AVOD \cite{wang2019pseudo} & - & 1.5 & 76.8 &65.1 & 56.6 & 60.7&- & 39.2 &-& 37.0& -\\
				PL++: AVOD \cite{journals/corr/abs-1906-06310} & - & 2.0 & 89.0 & 77.5  & 68.7 & 74.9 &66.83 & 56.8 &47.20& 49.0 & 40.30\\
				PL++: PIXOR \cite{journals/corr/abs-1906-06310} & - & 2.0 & 89.9 & 75.2  & 67.3 & 79.7 &70.7 & 61.1 &48.3& 54.5 & 41.0\\
				PL++: P-RCNN \cite{journals/corr/abs-1906-06310} & - & 2.0 & 88.4& 76.6 & 69.0  & 73.4 &- & 56.0 &-& 52.7 & -\\
				OC-Stereo \cite{journals/corr/abs-1909-07566}  & Titan Xp &2.9  & 90.01&80.63  & 71.06  & 77.66&68.89& 65.95 &51.47& 51.20& 42.97\\
				
				ZoomNet \cite{journals/corr/abs-2003-00529} & - & - & 90.62 & \textbf{88.40}  & \textbf{71.44}  & \textbf{78.68} &72.94 & \textbf{66.19} &\textbf{54.91}& \textbf{57.60} & \textbf{44.14}\\
				Disp R-CNN \cite{conf/cvpr/SunCXZJZB20} & - & 2.4 & \textbf{90.67} & 80.45  & 71.03 & 77.63 &\textbf{74.07} & 64.38 &52.34& 50.68 & 43.77\\				
				\hline
				\hline
				TL-Net \cite{conf/cvpr/0001CS19} & - & - & 62.46& 45.99  & 41.92  & 29.22 &- & 21.88 &-& 18.83 & - \\	
				Stereo RCNN \cite{conf/cvpr/0001CS19} & - & 2.4 & 87.13 & 74.11  & 58.93 & 68.50 &61.67 & 48.30 &43.87& 41.47 & 36.44\\				
				\hline
				Ours($iteration$=0, $resl=20$) & 2080Ti & 33.1& 89.88 & 78.05  & 69.17  & 73.43 &66.79 & 56.52 &45.22&48.29 & 38.48\\
				Ours($iteration$=1, $resl=20$) & 2080Ti & \textbf{\textcolor{blue}{25.4}}&\textcolor{blue}{90.58}\textcolor{red}{(-0.09)} &\textcolor{blue}{80.72}\textcolor{red}{(-7.68)}  & \textcolor{blue}{71.41}\textcolor{red}{(-0.03)}  & \textcolor{blue}{77.50}\textcolor{red}{(-1.18)} &72.17\textcolor{red}{(-1.9)} & \textcolor{blue}{58.65}\textcolor{red}{(-7.54)} &\textcolor{blue}{51.79}\textcolor{red}{(-3.13)}&\textcolor{blue}{50.14}\textcolor{red}{(-7.14)}  & \textcolor{blue}{43.19}\textcolor{red}{(-0.95)} \\
				\hline
		\end{tabular}}
		\label{tab:BEV}
	\end{center}
\end{table*}\\
\textbf{Structure-aware attention module.}
We first map the consistency of each voxel in FCE space to higher-dimensional vector $G_h \in \mathbb{R}^{1024\times resl\times resl\times resl}$ by point-wise multi-layer perceptron (MLP). Although the semantic-guided RBF can reduce the interference of non-target area noise, the FCE space still has a lot of spatial noise because the vehicle is a typical Non-Lambert. To address this issue, we present a StrAA module to reduce the interference of unstructured spatial noise. Many recent LiDAR-based methods \cite{8569311} can detect 3D objects on a bird‘s eye view (BEV), indicating that the top view contains the structural information needed for detection. Therefore, to determine if a particular point belongs to the structure of the object, we can search the boundary of the object on BEV from the average value of the height direction.  Specifically, StrAA first compute the average of $G_h$ in the height dimension as $G_{a} \in \mathbb{R}^{1024 \times resl \times resl} $, and then apply a standard 2D convolution with $3\times 3$ kernel size and $sigmoid$ to capture local structures. The output $G_{m} \in \mathbb{R}^{1024 \times resl \times resl} $ also can be regard as the attention map, inspired by self-attention \cite{vaswani2017attention}.
We obtain the final output $G_a \in \mathbb{R}^{1024 \times resl \times resl \times resl}$ by element-wise multiplication and summation. The overall process can be summarized as:
\begin{equation}
	\label{eq:a}
	\begin{aligned}
		G_a=\sigma \left(Conv^{3 \times 3}\left(Avg(G_h,dim=2)\right)\right) \otimes G_h + G_h
	\end{aligned}
\end{equation}
where $\otimes$ denotes element-wise multiplication. During multiplication, the attention map $G_m$ are broadcasted (copied) along the object hight dimension. After StrAA module, the $G_a$ are fed into the symmetric function following \cite{qi2016pointnet} to predict 3D box and its confidence.\\
\textbf{Losses for box prediction.} The box prediction head returns for each latent space with residual regression $\Delta B=(\Delta X, \Delta Y, \Delta Z, \Delta W, \Delta H, \Delta L, \Delta \theta)$ and its confidence $P_B$. Although these regression terms are independent, they are intrinsically related to the final box prediction. To sidestepping the issue of finding a proper weighting of each regression terms, we follow the $disentangling$ transformation\cite{simonelli2019disentangling} to decompose $\Delta B$ into 3 groups (dimensions $\phi_1$, position $\phi_2$, and orientation $\phi_3$) and unify the loss by the distance $L_{dis}$ of eight corners and one center between prediction and ground-truth. In short, the unify loss is computed as:
\begin{equation}
	\label{eq:a}
	\begin{aligned}
		L_{dis}(\phi,-\phi)=&\frac{1}{9}\sum\limits_{j=1}^{9}\left\|\pi(\Delta B(\phi,-\phi)+B_{ini}),\pi(B_{gt}) \right\|_2 \\
		&L_{reg}=\sum\limits_{m=1}^{3} L_{dis}(\phi_m,\phi_{-m}^{gt})
	\end{aligned}
\end{equation}
Here, $\pi:\mathbb{R}^7 \to \mathbb{R}^{3 \times 9}$ transform property of box to 3D coordinate of its eight corners and one center. $\phi_m$ denotes the sub-vector corresponding to the $m$th group, and $\phi_{-m}^{gt}$ denotes the sub-vector in ground truth corresponding to all but the $m$th group.

The confidence classification loss $L_{cls}$ aims to sort the quality of the target box. The label can be defined as:
\begin{equation}
	\label{eq:a}
	\hat{p}=
	\left\{
	\begin{array}{ccc}
		1       &      & IoU_{3D}>0.75\\
		0       &      &IoU_{3D}<0.25 \\
		2IoU_{3D}-0.5 &      & otherwise
	\end{array} \right.
\end{equation}
where $IoU_{3D}$ is the 3D intersection over union between prediction $\Delta B+B_{ini}$ and ground-truth $B_{gt}$. We then use the cross entropy loss to supervise the the predicted confidence. The overall training objective is:
\begin{equation}
	\label{eq:a}
	L=L_{reg}+\omega(t)L_{cls}
\end{equation}
Since the early training was unstable and the $IoU_{3D}$ was generally small, the time function $\omega(t)=exp[-5( 1-t/ 100)^2]$ was used to balance the weight of the two losses.
\section{Experimental}
\subsection{Implementation Details}
We evaluate the proposed approach on the KITTI 3D detection benchmark, which consists of 7481 training stereo images and 7518 test stereo images. We follow the protocol
in \cite{wang2019pseudo,conf/cvpr/SunCXZJZB20,journals/corr/abs-1909-07566} to split the training set as  $train$ set (3712 images) and $val$ set respectively,
and comprehensively compare proposed method with others on $val$ set as well as test set. We report two official evaluation metrics in KITTI: average precision for 3D detection ($AP_{3D}$) and bird's eye view detection ($AP_{BEV}$). We train our model on the machine E5-2678 CPU with two 2080Ti GPUs and apply $Adam$ optimizer with an initial learning rate of $0.000125$. We then train our model for 90 epochs and reduce the learning rate of $10 \times$ at $80$ epochs. Finally, $train$ set training takes 13 hours and the overall training set consumes 27 hours.

Establishing the FCE space needs the guidance of a coarse 3D box generated by the monocular-based methods. However, aligning the coarse 3D box to the ground truth is difficult. We, therefore, disturb the ground truth and then let our model predict this noise. We empirical set uniform noise in range $L=[-1.5, 1.5]$, $W=[-1.5, 1.5]$, $H=[-1.5, 1.5]$, $\theta =[-0.6, 0.6]$, $X=[-2, 2]$, $y=[-0.8, 0.8]$ and $Z=[-3, 3]$. 
\subsection{Comparison with Other Methods}
\begin{figure*}[htb]
	\begin{center}
		\includegraphics[width=2\columnwidth]{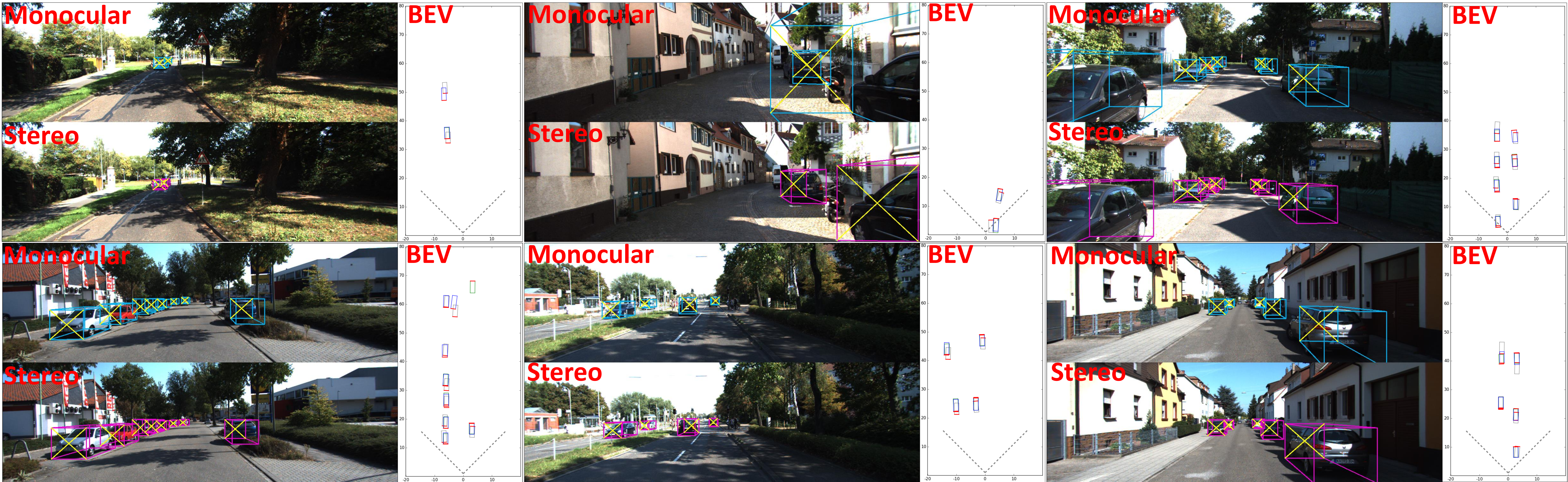}
	\end{center}
	\caption{Overview of the proposed 3D object detector. In BEV, Ground truth
		boxes are in \textcolor{green}{green}, stereo predicted boxes in \textcolor{blue}{blue} and monocular predicted boxes in \textcolor[rgb]{0.5,0.5,0.5}{gray}.}
	\label{fig:vis}
\end{figure*}

To fully evaluate the performance of the proposed method, we conduct our experiments in three regimes: easy, moderate, and hard, according to the occlusion and truncation levels. In addition to average precision, we also provide a comparison of runtime that is very important to the safety of autonomous driving or mobile robots. The results are shown in Table.\ref{tab:3d} and Table. \ref{tab:BEV}, we default use KM3D-Net to generate the initial latent space. We can observe that our RTS3D is the fastest running speed while our accuracy outperforms all image-only methods. Specifically, without extra labels helping, RTS3D is 10 times faster than the existing SOTA work Stereo RCNN while achieves $10\%$ improvement in $AP_{3D}$ and $AP_{BEV}$ for the moderate setting accuracy. Among other methods, the fastest is RT3DStereo, which requires depth and semantic masks for detection, while RTS3D only consumes $1/3$ of its runtime, and the accuracy in the easy set is increased by 105\%.
Moreover, compared with Pseudo-LiDAR, Pseudo-LiDAR++, DispRCNN, ZoomNet, and OC-Stereo, each of them needs to establish a 3D occupancy space or instance occupancy space with the help of a depth map, instance mask, or other labels, we can still obtain competitive detection accuracy but with minimal time consumption. We visualize some qualitative results of object detection in Fig. \ref{fig:vis}.
\subsection{Running-time Analysis}
In the case of a resolution $resl=20$, our RTS3D takes 5.6ms for the multi-scale feature extraction from left and right image, 21ms for the latent space generation, 7.6ms for the FCE space building, and 1.6ms for the 3D detection from the FCE space. The latent space generation and the multi-scale feature extraction can be executed in parallel and therefore the overall runtime is 30.2ms with iteration=1,
Note that these are the mean runtime over the $val$ set and can vary accordingly the number of the objects in stereo images.
\subsection{Ablation Study}
\begin{figure}[htb]
	\begin{center}
		\includegraphics[width=0.9\columnwidth]{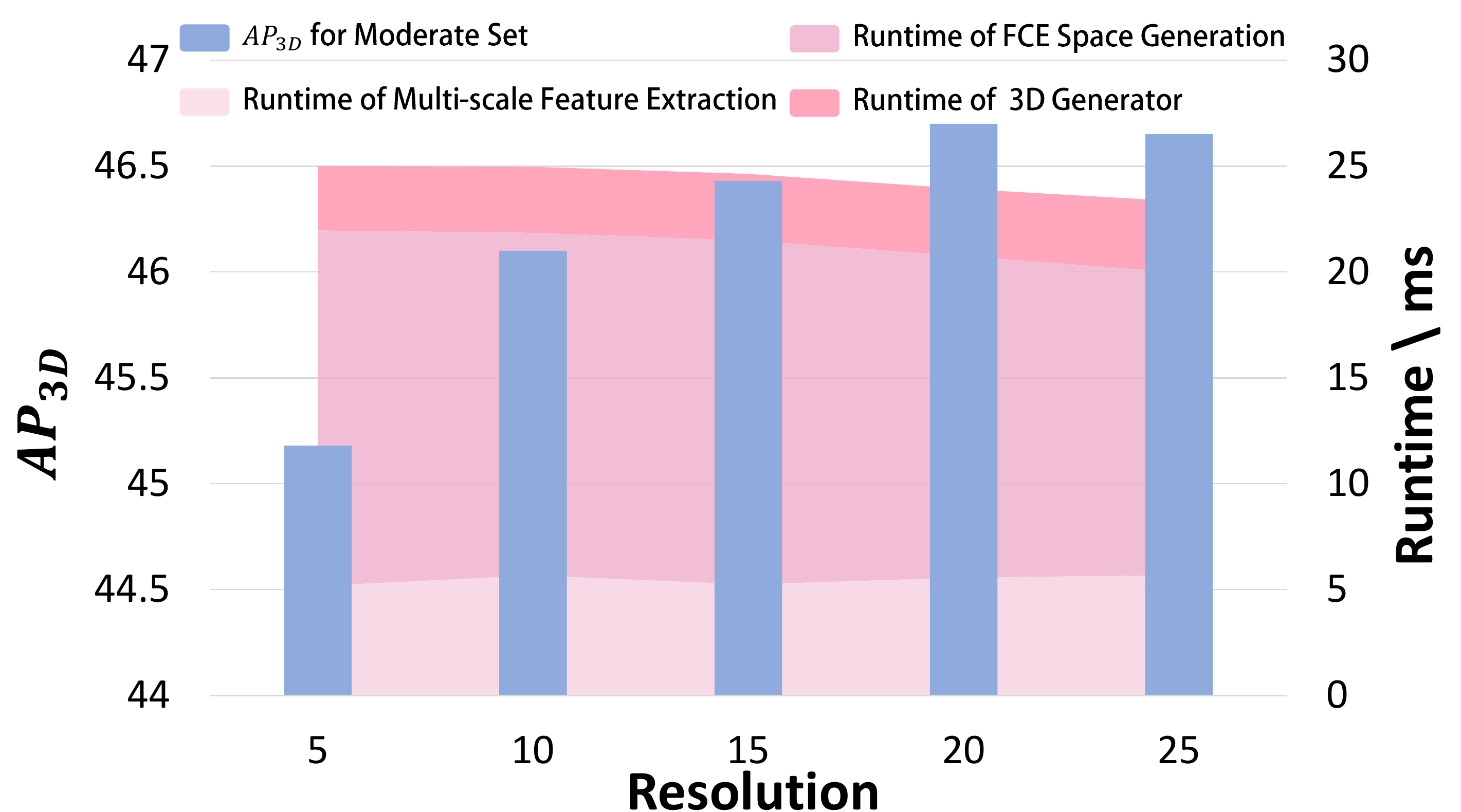}
	\end{center}
	\caption{Comparison of different resolutions with their runtime.}
	\label{fig:reslselect}
\end{figure}

\begin{figure}[htb]
	\begin{center}
		\includegraphics[width=0.9\columnwidth]{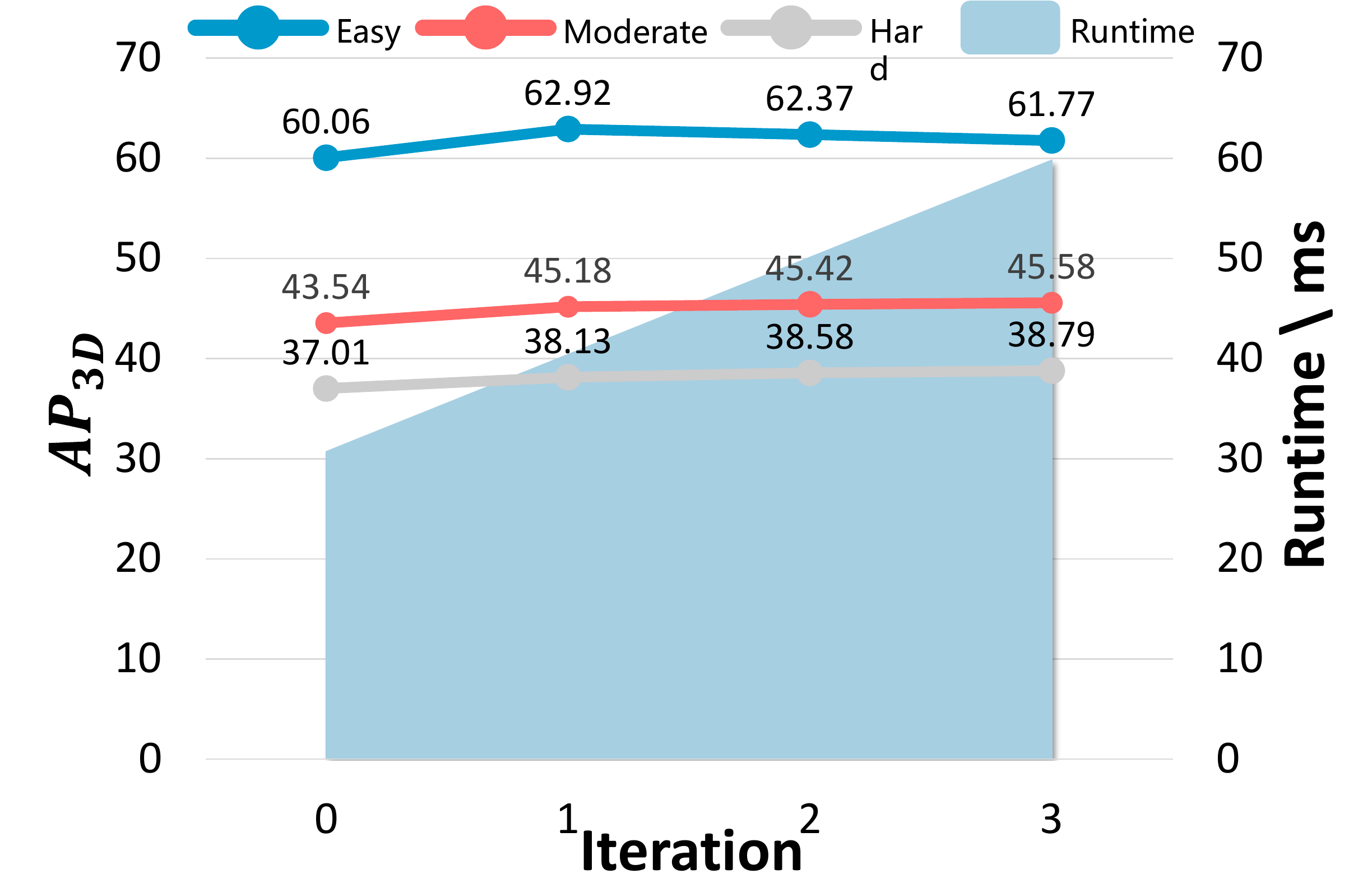}
	\end{center}
	\caption{Ablative Analysis of Iterations.}
	\label{fig:iteration}
\end{figure}

In this section, we perform comprehensive ablation experiments to validate the contributions of different components in our approach. All experiments are conducted on the $train$ split and evaluated on the $val$ split with the car category. If not specified, $resl$ for all experiments is set to $10$ and $interation$ for $1$.\\
\textbf{Feature-Consistency Embedding Space VS Occupancy Space.}
For comparing the FCE space and occupancy space, we first generate the 3D occupancy space by using the PSMNet\cite{conf/cvpr/ChangC18}.
We train the PSMNet\cite{conf/cvpr/ChangC18} on KITTI stereo ground truth to generate depth maps and transform this depth maps to 3D points cloud like most previous Pseudo-LiDAR similar works do \cite{wang2019pseudo,journals/corr/abs-1906-06310}. We then use 3D points in the latent space to train our 3D detectors. In this case, the StrAA module is not suitable for dealing with irregular point clouds. We, therefore, removed the StrAA module in all tests for a fair comparison. In addition, for the fairness of the time comparison experiment, The number of the input point cloud are sampled \cite{journals/cacm/Vitter84} to 1000 which is the same number as the voxel of our FCE space. The results are shown in Table \ref{tab:fcesVSos}. Only PSMNet alone is more computationally intensive than all pipeline of our method and it also need the supervision of ground truth that is usually generated by expensive LiDAR system. Finally, using our FCE space obtains strong improvement in both accuracy and running speed.
\begin{table}[htb]
	\caption{Comparisons of feature-consistency embedding space and occupancy space.}
	\footnotesize
	\begin{center}
		\begin{tabular}{c|c|c|c|c}
			\toprule[2pt]
			Config & Runtime & Set &{AP$_{\rm bev}^{0.7}$}  & {AP$_{\rm 3d}^{0.7}$}\\
			\hline
			\multirow{3}{*}{Occupancy Space} &  \multirow{3}{*}{418.4ms}& Easy & {52.46} & {34.93} \\
			\, &\,&  Mode  & {38.87} & {23.53} \\
			\, &\,&  Hard  &{33.27} & {21.15} \\
			\hline
			
			\multirow{3}{*}{FCE Space} &  \multirow{3}{*}{39.4ms}&Easy & {74.12} & {60.80} \\
			\, &\,&  Mode  & {56.08} & {43.54} \\
			\, &\,&  Hard  & {47.33} & {35.70} \\
			\hline
			\multirow{3}{*}{FCE Space with StrAA} &  \multirow{3}{*}{39.4ms}&Easy & {76.29} &{62.92} \\
			\, &\,&  Mode  & {57.58} & {45.18} \\
			\, &\,&  Hard  & {48.99} & {38.13} \\
			\bottomrule[2pt]
		\end{tabular}
		\label{tab:fcesVSos}
	\end{center}
\end{table}
\begin{table}[htb]
	\scriptsize
	\caption{Ablative analysis of the different methods for generating the feature-consistency space. Only the moderate sets are reported.}
	\begin{center}
		\begin{tabular}{c|c|c|c}
			\toprule[2pt]
			Config & Method& {AP$_{\rm bev}^{0.7}$}  & {AP$_{\rm 3d}^{0.7}$}\\
			\hline
			\multirow{4}{*}{Channel-reducing}  & Concatenation with MLP & 15.74  & 11.21 \\
			\,                                & Cosine Correlation     & 36.38  & 21.73 \\
			\,                               &Gaussion                & 38.87  & \textbf{23.53} \\
			\,                                & RBF with MLP           &\textbf{39.39}  & 21.15\\
			\hline
			\multirow{3}{*}{Channel-keeping}   & Concatenation with MLP & 20.62  & 13.03 \\
			\,                                & Absolute Difference     & 53.08  & 39.54 \\
			\,                                & RBF                    & \textbf{57.18}  & \textbf{45.18} \\
			\bottomrule[2pt]
		\end{tabular}
		\label{tab:RBF}
	\end{center}
\end{table}\\
\textbf{Structure-Aware Attention Module.}
We further evaluate the effect of the proposed StrAA module.
The results are shown in Table \ref{tab:fcesVSos}. Without bells and whistles, we already outperform most of the image-based methods in both accuracy and running speed. StrAA module further enhances our method performance by several points in a slight computation increasing.\\
\textbf{Semantic-guided RBF.}
Table \ref{tab:RBF} compares different types of generating feature-consistency space. With semantic-guided RBF, the proposed method obtains a better detection accuracy than the absolute difference. In channel-keeping methods, concatenation has the worst results. A possible reason is that the implicit model is difficult to learn effective knowledge without point-wise supervision. Channel-reducing methods average the channel for forcing the model to predict the occupancy of a point. All these methods have a poor performance, which also demonstrates the advantage of learning from the original image feature space.
\textbf{Resolution of Feature-Consistency Embedding Space.}
We examine the accuracy and running speed of our method with respect to the resolution of the FCE space. The results are shown in Fig. \ref{fig:reslselect}. We observe that the accuracy will increase as the resolution of the FCE space increase until it reaches $20$. A larger resolution will bring more details, but a small $batch size$ will also cause training instability. It is worth noting that even with a very small resolution, we still obtain a relatively good detection accuracy and have extremely fast running speed.
\\
\textbf{Different Monocular 3D Detectors.}
We compare the impact of different monocular detectors on the generation of initial latent space. As shown in Table \ref{tab:dm3d}, even if the accuracy of the monocular detection method varies greatly, the final stereo accuracy is similar.
This is likely due to two reasons. First, we use the iterative method to continuously modify this initial latent space. Second, RTS3D is trained on pseudo data, so it is not sensitive to the initial latent space generated by different monocular methods.
\begin{table}[htb]
	\caption{Ablative analysis of different methods to generate initial latent space. }
	\begin{center}
		\tiny
		\begin{tabular}{c|c|c|c|c|c|c|c}
			\toprule[2pt]
			\multirow{2}{*} {Config}&  \multirow{2}{*} {Data} & \multicolumn{3}{c|}{$AP_{BEV}^{0.5}$}   &  \multicolumn{3}{c}{$AP_{3D}^{0.5}$}  \\
			\cline{3-8}
			&  & Easy&Moderate&Hard&Easy&Moderate&Hard\\
			\hline
			\multirow{2}{*} {CenterNet}&Mono &27.22&22.91&19.52&13.53&10.37&10.56\\
			&Stereo &\textbf{90.03}&\textbf{77.69}&\textbf{70.39}& \textbf{89.32}& \textbf{78.18}&\textbf{68.13}\\
			\hline
			\multirow{2}{*} {KMNet}&Mono &53.77&40.58&34.79&47.23&34.12&31.51\\
			&Stereo &\textbf{90.44}&\textbf{79.98}&\textbf{70.75}& \textbf{90.27}& \textbf{78.27}&\textbf{69.06}\\
			\hline
		\end{tabular}
		\label{tab:dm3d}
	\end{center}
\end{table}\\
\textbf{Ablative Analysis of Iterations}
We also compare the effects of different iterations on accuracy and runtime, as shown in Fig. \ref{fig:iteration}. When iterations exceed 1, the effect
is not significantly improved, but the running time will be greatly increased. Therefore, the number of iterations in our experiment is default set as 1 for the best speed-accuracy trade-off.
\\
\section{Conclusion and Discussion.}
We present a novel framework to perform faster and more accurate 3D object detection using stereo images. We design a novel 3D intermediate representation space which can encode the structural and semantic information of object without relying on additional annotation. We then propose a semantic-guided RBF and structure-aware attention module for reducing the influence of space noise. Extensive experiments show that our model achieves an unprecedented running speed while competing with the most advanced methods for accuracy.

Exploring the intermediate representation of a 3D scene has always been a meaningful thing. Pseudo-LiDAR transforms a front-view image with estimated depth map to 3D occupancy representation, bridging the gap between the LiDAR- and image-based detection accuracy. We propose a 4D feature-consistency representation to further bridge this gap and greatly improve the detection speed. We believe that the rapid progress in speed can not only greatly ensure the safety of autonomous driving, but also can further enhance accuracy in additional ways. One of the most straightforward methods conceivable is to smooth the detection results between adjacent frames. This is what we will do in future work.
\bibliography{ref}
\end{document}